\documentclass[runningheads]{llncs}
\usepackage{graphicx}
\usepackage{wrapfig,lipsum}
\usepackage{url}
\usepackage[hidelinks]{hyperref}
\usepackage[utf8]{inputenc}
\usepackage[small]{caption}

\usepackage{amsmath}
\usepackage{graphicx}
\usepackage{booktabs}
\usepackage{algorithm}
\usepackage{algorithmic}
\usepackage{amsfonts}
\usepackage{amssymb}
\usepackage{color}
\usepackage{multirow}
\usepackage[misc]{ifsym}
\begin{document}
\title{Supervised Graph Contrastive Learning for Few-shot Node Classification}
%
%\titlerunning{Abbreviated paper title}
% If the paper title is too long for the running head, you can set
% an abbreviated paper title here
%
% \author{Anonymous authors}
\author{Zhen Tan \Letter\inst{1} \and
Kaize Ding\inst{1} \and
Ruocheng Guo\inst{2} \and
Huan Liu\inst{1}}

\authorrunning{Z. Tan et al.}
% First names are abbreviated in the running head.
% If there are more than two authors, 'et al.' is used.
%
% \institute{Anonymous institues}
\institute{Arizona State University, Tempe AZ, USA
\and
Bytedance AI Lab, London, UK
\\
\email{\{ztan36,kding9,huanliu\}@asu.edu}, \email{rguo.asu@gmail.com}
}

% Springer Heidelberg, Tiergartenstr. 17, 69121 Heidelberg, Germany

% \url{http://www.springer.com/gp/computer-science/lncs} \and
% ABC Institute, Rupert-Karls-University Heidelberg, Heidelberg, Germany\\
% \email{\{abc,lncs\}@uni-heidelberg.de}}
%
\maketitle              % typeset the header of the contribution
\begin{abstract}

Graphs present in many real-world applications, such as financial fraud detection, commercial recommendation, and social network analysis. But given the high cost of graph annotation or labeling, we face a severe graph label-scarcity problem, i.e., a graph might have a few labeled nodes. One example of such a problem is the so-called \textit{few-shot node classification}. A predominant approach to this problem resorts to \textit{episodic meta-learning}. In this work, we challenge the status quo by asking a fundamental question whether meta-learning is a must for few-shot node classification tasks. We propose a new and simple framework under the standard few-shot node classification setting as an alternative to meta-learning to learn an effective graph encoder. The framework consists of supervised graph contrastive learning with novel mechanisms for data augmentation, subgraph encoding, and multi-scale contrast on graphs. Extensive experiments on three benchmark datasets (CoraFull, Reddit, Ogbn) show that the new framework significantly outperforms state-of-the-art meta-learning based methods.

\end{abstract}
\section{Introduction}

%Due to the strong modeling capability, a spectrum of 
Graphs are ubiquitous in many real-world applications. Graph Neural Networks (GNNs)~\cite{kipf2016semi,velivckovic2017graph,xu2018powerful} have been applied to model a myriad of network-based systems in various domains, such as social networks \cite{hamilton2017inductive}, citation networks \cite{jiao2020sub}, and knowledge graphs~\cite{park2019estimating}. %Following the recurrent mechanism of feature transformation and propagation, GNNs succeed in learning expressive representations to solve problems in all those domains.
Despite these breakthroughs, it has been noticed that conventional GNNs fail to make accurate predictions when the labels are scarcely available~\cite{ding2020graph,zhang2022few,ding2022meta}. One such problem is so-called \textit{few-shot node classification}. It consists of two disjoint phases: In the first phase (train),  classes with substantial labeled nodes (i.e., \textit{base classes}) are available to learn a GNN model; and in the second phase (test), the GNN classifies nodes of unseen or \textit{novel classes} with few labeled nodes. A few-shot node classification task is called \textit{$N$-way $K$-shot node classification} if a node is classified into $N$ classes and each class contains only a few $K$ labeled nodes in the test phase. 

\begin{figure}[h]
  \centering
  \scalebox{0.9}{
  \includegraphics[width=\linewidth]{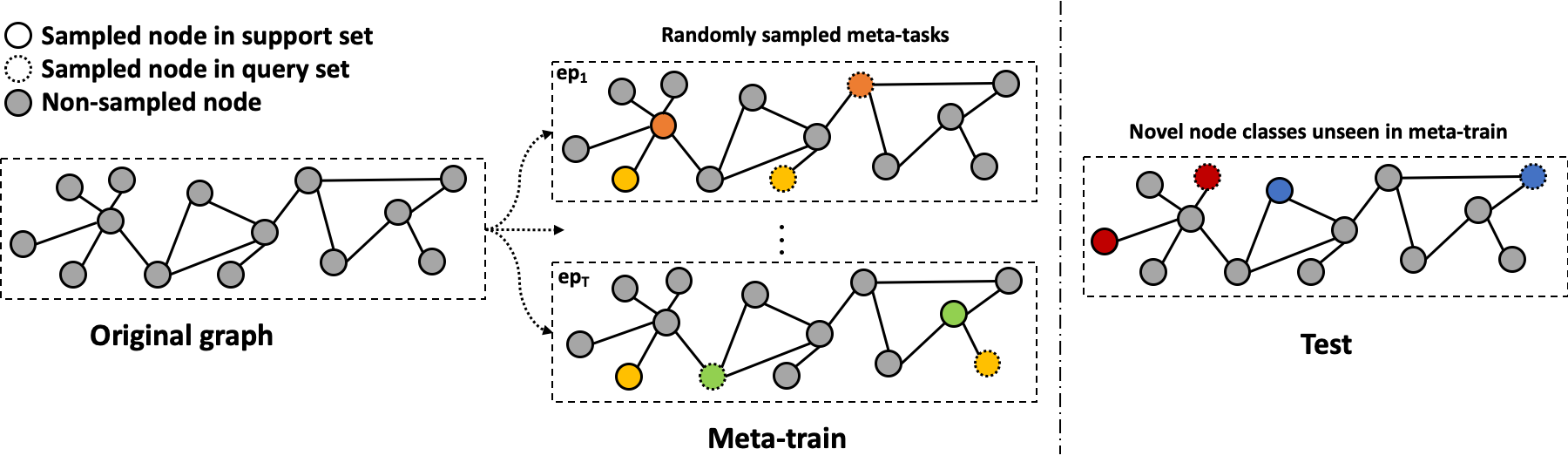}}
  \vspace{-0.1cm}
  \caption{Episodic meta-learning for few-shot node classification (ep$_i$ is the $i$th episode). Colors indicate different classes. Specially, grey nodes mean those nodes are not sampled. Different types of nodes indicate if nodes are from a support set or a query set.}
  \label{fig:meta}
  \vspace{-0.4cm}
\end{figure}

This shortage of labeled training data for the novel classes poses a great challenge to learning effective GNNs. A prevailing paradigm to tackle this problem is \textit{episodic meta-learning} \cite{zhou2019meta,ding2020graph,wang2020graph,guo2021few,liu2021relative}; its representative algorithms are Matching Network~\cite{vinyals2016matching}, MAML~\cite{finn2017model}, and Prototypical Network~\cite{snell2017prototypical}. Episodic meta-learning is inspired by how humans learn unseen classes with few samples via utilizing previously learned prior knowledge. During the training phase, it generates numerous meta-train tasks (or episodes) by emulating the test tasks, following the same $N$-way $K$-shot node classification structure. An example of episodic meta-learning is shown in Figure~\ref{fig:meta}. In each episode, $K$ labeled nodes are randomly sampled from $N$ base classes, forming a \textit{support set}, to train the GNN model while emulating the $N$-way $K$-shot node classification in the test phase. The GNN then predicts labels for an emulated \textit{query set} of nodes randomly sampled from the same classes as the support set. A Cross-Entropy Loss is used for backpropagation to update the GNN. 
%Usually, to better capture the topological relationship within the whole graph from the sampled pieces, extra sophisticated modules are involved \cite{ding2020graph,wang2020graph,liu2021relative}. 
Current research~\cite{zhou2019meta,ding2020graph,wang2020graph,guo2021few,ding2021few} has shown that via numerous such episodic emulations across different sampled meta-tasks on bases classes, the trained encoder can extract transferable meta-knowledge to fast adapt to unseen novel classes.
\vspace{-0.4cm}

%However, in this paper, we question the necessity of meta-learning for few-shot node classification. 
These episodic meta-learning based methods entail the following steps: (1)~random sampling in each episode for meta-train in order to acquire the topological knowledge crucial for learning representative node embeddings. Since only a small portion of the nodes and classes are randomly selected per episode, the topological knowledge learned with those nodes is piecemeal and insufficient to train expressive GNN encoder, especially if the selected nodes share little correlation. (2) To boost accuracy, those meta-learning based methods have to rely on a large number of episodes. % or complex optimization mechanisms. 
In other words, a large number of samples from the original graph is required for meta-train to capture the topological knowledge that can be transferred for use in the test phase. Consequently, these meta-learning algorithms can take time to converge with their emulation-based meta-train. The two problems are closely related. The \textit{piecemeal knowledge} from emulation-based meta-learning unique for few-shot node classification entails the need for \textit{large numbers of episodes} to acquire better topological knowledge. In this paper, we investigate if an alternative approach to episodic meta-learning can be developed so that the two problems can be addressed from their root causes for better performance of few-shot node classification. We posit that the key to few-shot node classification is to learn a generalizable GNN encoder that can produce discriminative representation on novel classes by learning transferable topological patterns implied in bases classes. If we can address the piecemeal knowledge problem, we can better use the few labeled nodes from novel classes to fine-tune another simple classifier (e.g., Logistic Regression, SVM, shallow MLP, etc.) to predict labels for other unlabeled nodes. %as shown in Figure \ref{fig:meta} (b). 

As an alternative to episodic meta-learning, we propose a novel approach for few-shot node classification, supervised graph contrastive learning~\cite{khosla2020supervised,gunel2020supervised}. Graph contrastive learning is proven effective in training powerful GNNs~\cite{you2020graph,hassani2020contrastive,zhu2021graph}. %To generate transferable and discriminative node representations, our proposed approach augments data using nodes and subgraphs to pretrain a GNN encoder.  
Multiple views of the original graph are first created through predefined transformations~\cite{you2020graph,hassani2020contrastive,zhu2021graph,ding2022data} (e.g., randomly dropping edges, randomly perturbing node attributes). Then a contrastive loss is applied to maximize feature consistency under differently augmented views. However, none of those existing methods accommodate the unique characteristics of few-shot node classification. In this paper, we propose a novel graph contrastive learning method especially designed for few-shot node classification. We will present technical details on how the new supervised contrastive learning can avoid the two problems with the episodic meta-learning after a formal problem statement is given.

\paragraph{Contributions.} Our contributions include: (1) we are the first to investigate an alternative to the prevailing meta-learning paradigm for graph few-shot learning; (2) we propose a supervised graph contrastive learning method tailored for few-shot node classification by developing novel mechanisms for data augmentation, subgraph encoding, and multi-scale contrast on a graph; and (3) we conduct systematic experiments to assess the proposed framework  in comparison with the existing meta-learning based graph few-shot learning methods and representative graph contrastive learning methods in terms of accuracy and efficiency. 

\section{Problem Formulation}

In this paper, we focus on few-shot node classification on a single graph. Formally, given an attributed network $G = (\mathcal{V},\mathcal{E}, \textbf{X}) = (\textbf{A}, \textbf{X})$, where $\mathcal{V}$, $\mathcal{E}$,  $\textbf{A}$ and $\textbf{X}$ denote the nodes, edges, adjacency matrix and node attributes, respectively. The few-shot node classification problem assumes the existence of a series of homogeneous node classification tasks $\mathcal{T} = \{\mathcal{D}^i\}^{I}_{i=1} = \{(\textbf{A}_{\mathcal{C}^i}, \textbf{X}_{\mathcal{C}^i})\}^I_{i=1}$, where $\mathcal{D}^i$ denotes the given dataset of a task, $I$ denotes the number of such tasks, $\textbf{X}_{\mathcal{C}^{i}}$ denotes the attributes of nodes whose labels belong to the label space $\mathcal{C}^{i}$, and $\textbf{A}_{\mathcal{C}^{i}}$ similarly. Following the literature \cite{zhou2019meta,ding2020graph,wang2020graph,guo2021few,liu2021relative}, we call the classes available during training as base classes, $\mathcal{C}_{base}$, and the classes for target test phase as novel classes, $\mathcal{C}_{novel}$, $\mathcal{C}_{base} \cap \mathcal{C}_{novel} = \varnothing$. Conventionally, there are substantial gold-labeled nodes for $\mathcal{C}_{base}$, but few labeled nodes for novel classes $\mathcal{C}_{novel}$. We can formally define the problem of few-shot node classification as follows:

\begin{definition}
\textbf{Few-shot Node Classification:} Given an attributed graph $G = (\textbf{A}, \textbf{X})$ with a divided node label space $\mathcal{C} = \{\mathcal{C}_{base}, \mathcal{C}_{novel}\}$, we have substantial labeled nodes from $\mathcal{C}_{base}$, and few-shot labeled nodes (support set $\mathcal{S}$) for $\mathcal{C}_{novel}$. The task is to predict the labels for unlabeled nodes (query set $\mathcal{Q}$) from $\mathcal{C}_{novel}$. %If each support set has $N$ novel classes with $K$ labeled nodes, then we term this task an $N$-way $K$-shot node classification task.
\end{definition}

\section{Methodology}
% Due to the high dimensionality of the node attributes and the limited number of labeled nodes, input nodes are usually replaced by their features produced by a graph encoder $g_\theta$ (parameterized by $\theta$). 
In graph representation learning, usually a GNN encoder $g_\theta$ is employed to model the high dimensional graph knowledge and project the node attributes to a low dimensional latent space. The classifier $f_\psi$ is then applied to the latent node representations for node classification. The essence of few-shot node classification is to learn a encoder $g_\theta$ that can transfer the topological and semantic knowledge learned from substantial data of base classes to generate discriminative embeddings for nodes from novel classes with limited supervisory information.

As a prevailing paradigm, meta-learning is adopted~\cite{zhou2019meta,ding2020graph,wang2020graph} to jointly learn $g_\theta$ and $f_\psi$ by episodically optimizing the Cross-Entropy Loss  (CEL) on sampled meta-tasks. However, optimizing CEL over sampled piecemeal graph structures will engender node embeddings excessively discriminative against the current sampled nodes, rendering them sub-optimal for nodes classification in the test phase where the nodes are sampled arbitrarily from unseen novel classes. To mitigate such issues, those meta-learning based methods rely on a large number of episodes to train the model on numerous differently sampled meta-tasks to learn more transferable node embeddings, which makes the training process highly unscalable, especially for large graphs. 

\begin{figure}[h]
  \centering
  \includegraphics[width=\linewidth]{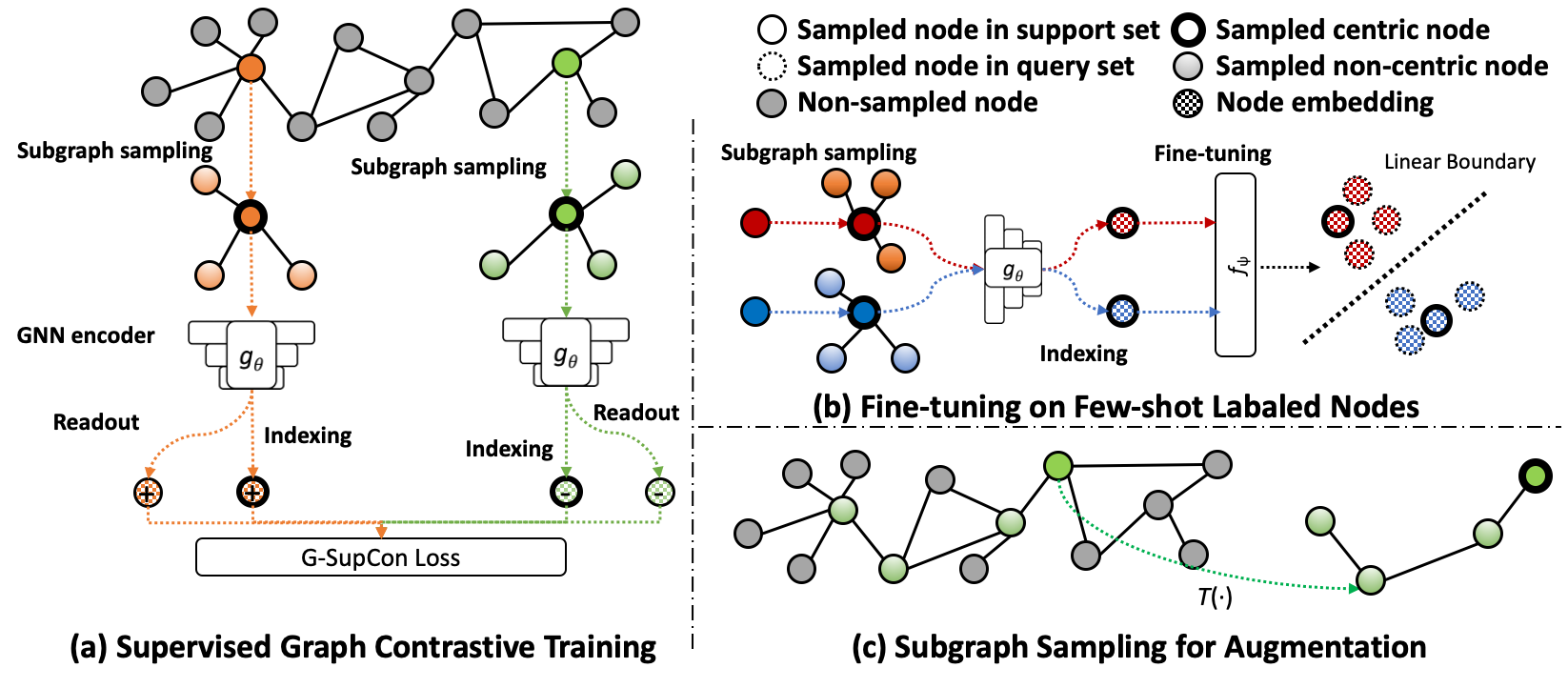}
  \caption{Colors indicate different classes. Specially, grey nodes mean those nodes are not sampled. Different types of nodes indicate if nodes are from a support set or a query set, or sampled as centric nodes. Ombré nodes indicate the nodes are sampled non-centric nodes for the subgraphs. Crisscross nodes are node embeddings from the GNN encoder. (a) Supervised graph contrastive training framework. (b) Fine-tuning on few-shot labeled nodes from novel classes. (c) Node connectivity based subgraph sampling strategy samples nodes that are strongly connected to the centric nodes.}
  \label{fig:ft}
  \vspace{-0.0cm}
\end{figure}

As a remedy, in this paper, we propose a decoupled method to learn the graph encoder $g_\theta$ and the final node classifier $f_\psi$ separately. We put forward a supervised graph contrastive learning to firstly pretrain the GNN encoder $g_\theta$ to generate more transferable node embeddings. With such high-quality node embeddings, we can fine-tune a simple linear classifier to perform the final few-shot node classification. As shown in Figure \ref{fig:ft}, our framework consists of the following key components:
\begin{itemize}
\item An augmentation function $T(\cdot)$ that transforms a sampled centric node into a correlated view. We propose a node connectivity based augmentation mechanism to sample nodes that are highly correlated to the centric nodes to form a subgraph as its augmented view (Section \ref{da}).
\item A GNN encoder $g_\theta$ that encodes the subgraphs rather than the whole graph like all the existing meta-learning methods do. In such a manner, our model will consume much less time to converge (Section \ref{se}).

\item A contrastive mechanism that enables the encoder $g_\theta$ to discriminates embeddings between differently augmented views by capturing structural patterns across base classes and extrapolating such knowledge onto unseen novel classes (Section \ref{clav}). 
\item A linear classifier $f_\psi$ fine-tuned by a few labeled nodes from novel classes and is tasked to predict labels for those unlabeled nodes (Section \ref{lcft}).
\end{itemize}

\subsection{Data Augmentation}
\label{da}
Given a graph $\mathcal{G} = (\mathcal{V},\mathcal{E}, \textbf{X}) = (\textbf{A}, \textbf{X})$, following the conventions in contrastive learning methods \cite{chen2020simple,he2020momentum}, a transformation $T(\cdot)$ is used to generate a new view $\textbf{x}^\prime_j$ for a node $j$ ($\forall j \in \{1,...,M\}$, $M$ is the number of nodes) with node attributes $\textbf{X}$ and the adjacency matrix $\textbf{A}$:
\begin{equation}
    \textbf{x}^{\prime}_j = T(\textbf{A}, \textbf{X}, j)
\end{equation}
There are multiple transformations for graph data, such as node masking or feature perturbation. However, such methods can introduce extra noise that impairs the learned node representations. In this paper, we propose a node connectivity based subgraph sampling strategy as the data augmentation mechanism. Connectivity score is a family of metrics that measures the connection strength between a pair of nodes in a graph without using the node attributes \cite{chen2009measure,jiao2020sub}. Notably, since only the adjacency matrix is needed for calculation, such augmentation can be pre-computed before training:
\begin{equation}
    \textbf{S} = Connect(\mathcal{V}, \mathcal{E}) = Connect(\textbf{A}),
\end{equation}
where each column $\textbf{s}_j$ of $\textbf{S}$ contains the scores between node $j$ and all nodes in the graph. We set the score between a node and itself as a constant value $\gamma = 0.3$ for better performance. Intuitively, nodes that share more semantic similarities tend to have more correlations. However, such nodes may not be geologically close to each other. Node connectivity can capture the correlation between nodes by considering both the global and local graph structures \cite{chen2009measure,jeh2003scaling}. As shown in Figure \ref{fig:ft} (c), for any given node, we treat it as the centric node and sample other nodes that have the highest connectivity scores with the centric node to build a contextualized subgraph as its augmented view. Now the transformation for data augmentation can be defined as:
\begin{equation}
\begin{split}
    \textbf{x}^{\prime}_j &= T(\textbf{A}, \textbf{X}, j) \\
    &= \textbf{X}[top\_rank(\textbf{S}[j, :], \alpha)] \\
    &= \textbf{X}[top\_rank(Connect(\textbf{A})[j, :], \alpha)] \\
    &= (\textbf{A}^\prime_j, \textbf{X}^\prime_j) =  \mathcal{G}_s(j) 
\end{split}
\end{equation}
where $top\_rank$ is a function that returns the indices of the top $\alpha$ values, $\alpha$ is a hyperparamter that controls the augmented subgraph size ($\alpha + 1$), and $\mathcal{G}_s(j)$ is the augmented subgraph for node $j$ with adjcency matrix $\textbf{A}^\prime_j$ and sampled node attributes $\textbf{X}^\prime_j$. In this paper, we present two methods to calculate the node connectivity scores: Node Algebraic Distance (NAD) \cite{chen2009measure} and Personalized PageRank (PPR)~\cite{jeh2003scaling}.

\subsubsection{Node Algebraic Distance (NAD).}
Following \cite{chen2009measure}, we first randomly assign a random value $u_i$ $(0 < u_i < 1)$ to any node $i$ in the graph to form a vector $\textbf{u} \in \mathbb{R}^{M}$. Then, we iteratively updates the value of a node by aggregating its neighboring weighted values:
At the $t$-th iteration, for node $i$ we have:
\begin{equation}
    \hat{u}_i^{t} = \sum_j \textbf{A}(i,j)u_j^{t-1}/\sum_j \textbf{A}(i,j),
\end{equation}
\begin{equation}
    \textbf{u}^{t} = (1-\eta)\textbf{u}^{t-1} + \eta  \hat{\textbf{u}}^{t},
\end{equation}
where $\eta$ is a parameter set to $0.5$. After a few iterations, the difference between the values of node $i$ and $j$ indicates the coupling between them. The smaller difference stands for a stronger connection. The final score matrix is:
\begin{equation}
    \textbf{S} = \{\textbf{S}(i,j)\}_{i,j=1}^{M} = \frac{1}{|u_i - u_j| + \epsilon}
\end{equation}
where $\epsilon$ is a parameter set to $0.01$. We column-normalize $\textbf{S}$ and then set the value of $\textbf{S}(i,i), (\forall i \in M)$ to $\gamma$.

\subsubsection{Personalized PageRank (PPR).}
For PPR we have:
\begin{equation}
    \mathbf{S} = \phi \cdot (\textbf{I} - (1-\phi) \cdot \textbf{A}\textbf{D}^{-1})
\end{equation}
where $\phi$ is a parameter usually set as $0.15$, and $\textbf{D}$ is a diagonal matrix with: $\textbf{D}(i,i) = \sum_j\textbf{A}(i,j)$. We column-normalize the scores $\textbf{S}$ to make the scores on the diagonal equal to $\gamma = 0.3$ as in NAD.

\subsection{Subgraph Encoder}
\label{se}
With the pre-computed connectivity score matrix, for each node $j$, we sample the top $\alpha$ nodes with the highest connectivity scores to construct a subgraph $\mathcal{G}_s(j) = (\textbf{A}^\prime_j, \textbf{X}^\prime_j)$. The larger $\alpha$ is, the richer context is given in the subgraph (We set $\alpha = 19$ for memory limitation). Then, as shown in Figure \ref{fig:ft} (a), we feed the resulting subgraph into a GNN encoder $g_\theta$. \label{fast} In particular, as the size of the subgraphs is a fixed small number (20 in our case) as opposed to the original graph, compared to the existing methods where the encoder encodes the whole graph, our method reduces the dimension for neighborhood aggregation to a much smaller magnitude and makes it highly scalable and faster to converge:
\begin{equation}
    \textbf{Z}^{\prime}_j = g_\theta(\textbf{x}_j^\prime) = g_\theta(\textbf{A}^\prime_j, \textbf{X}^\prime_j),
\end{equation}
where $\textbf{Z}^\prime_j \in \mathbb{R}^{(\alpha + 1)\times F}$ and $F$ is the embedding size. We normalize the latent features for better performance. To enable the contrastive learning across different scales (e.g. nodes and subgraphs), we propose a readout function $R(\cdot)$ which maps the representation of all nodes in the subgraph, $\textbf{Z}^\prime_j$, to a vector $\textbf{z}^\prime_j$ as the summarized subgraph representation. Intuitively, $\textbf{z}^\prime_j$ is a weighted average of the representations of all nodes in the subgraph. The weights are their normalized connectivity scores $\hat{\textbf{s}}_j \in \mathbb{R}^{\alpha}$ to node $j$. We set the connectivity score of a node $j$ to itself as a constant value $\gamma$ (as defined in Section \ref{da}) before normalization:
\begin{equation}
    \textbf{z}^{\prime}_j = R(\textbf{Z}^\prime, \hat{\textbf{s}}_j) = \sigma(\hat{\textbf{s}}_j^T \cdot \textbf{Z}^\prime )
\end{equation}
where $\sigma$ is the sigmoid function, and $\textbf{z}^\prime_j \in \mathbb{R}^{F}$ is the subgraph representation. The centric node embedding $\textbf{z}_j\in \mathbb{R}^{F}$ can be directly indexed from $\textbf{Z}^\prime_j$.

\subsection{Multi-scale Graph Contrastive Learning with Augmented Views}
\label{clav}

Since the augmented subgraphs contain topological information crucial for learning expressive encoder, we propose to use multi-scale contrastive learning that combines three categories of contrastive pairs to enforce the model to learn from both individual attributes and contextualized knowledge from sampled views: the contrast between (1) nodes and nodes ($\textbf{z}_i \,\&\, \textbf{z}_j$), (2) subgraphs and subgraphs ($\textbf{z}_i^\prime \,\&\, \textbf{z}_j^\prime$), and (3) nodes and subgraphs ($\textbf{z}_i \,\&\, \textbf{z}_j^\prime$), where $i$, $j$ are arbitrary node indices. For any node, the nodes in the same class together with their corresponding subgraphs are viewed as positives, and all the rest are viewed as negatives. Now, given the batch size $B$, we define a \textit{duo-viewed} batch, consisting of the representations of both nodes and their corresponding augmented subgraphs:
\begin{equation}
\{(\textbf{h}_b, y_b)\}_b^{2B} = \{(\textbf{z}^\prime_b, y_b), (\textbf{z}_b, y_b) \}_b^{B}.
\end{equation}

\subsubsection{Loss Function}

To better suit the setting of few-shot node classification, we adapt the Supervised Contrastive loss (SupCon) from \cite{khosla2020supervised} to pretrain the GNN encoders. We term the proposed loss G-SupCon for convenience. Compared to unsupervised contrastive loss (e.g. Deep InfoMax (DIM) \cite{bachman2019learning}, SimCLR \cite{chen2020simple}, Margin Loss~\cite{jiao2020sub} etc.), and Cross-Entropy Loss (CEL), G-SupCon utilizes the ground-truth label to sample mini-batches to help to better align the representation of nodes in the same class more closely and push nodes from different classes further apart. Such learning patterns can be easier to transfer to unseen novel classes to generate highly discriminative representations. To ensure the balance in training data, we sample $B/|C_{tr}|$ nodes per class as centric nodes in each mini-batch for training, where $|C_{tr}|$ is the number of classes for pretraining (i.e., base classes). We term it Balanced Sampling (BS) \label{BS} for convenience. Then, the loss function is defined as:
\begin{equation}
    \mathcal{L} = \sum_{b \in B}\frac{-1}{|P(b)|}
    \sum_{p \in P(b)} \log {\frac{\exp{(\textbf{h}_b \cdot \textbf{h}_p/\tau)}}{\sum_{a \in A(b)} \exp{(\textbf{h}_b \cdot \textbf{h}_a/\tau)}}},
\end{equation}
where $A(b)$ is the set of indices from 1 to $2B$ excluding b, and $P(b)$ is the set of indices of all positives in a duo-viewed batch excluding $b$. This contrastive loss includes all the three categories of contrastive pairs mentioned before. Additionally, $\tau \in \mathbb{R}^+$ is a scalar temperature parameter defined as $\tau = \beta/\sqrt{degree(\mathcal{G})}$, where $degree(\mathcal{G})$ is the average degree of the graph, and $\beta$ is a hyperparameter. The temperature parameter controls the sensitivity of the trained model to the hard negative samples. As nodes in different classes also share some correlation with the centric nodes, especially for more complex graphs with higher average degrees, we do not want them to be fully separated apart in the representation space. So, we add the degree as a penalty for the temperature parameter to guide the separateness. 

Under this fully-supervised setting, we pretrain our GNN encoder with all the base node labels and fix it during fine-tuning.

\subsection{Linear Classifier Fine-tuning}
\label{lcft}
As indicated in Figure \ref{fig:ft} (b), with a pretrained GNN encoder, when fine-tuning on a target few-shot node classification dataset $\mathcal{D}^i$ on novel classes, we tune a separate linear classifier $f_\psi$, (e.g. logistic regression, SVM, a linear layer, etc.) with the few labeled nodes in the support set $\mathcal{S}^i$, and task it to predict the labels for nodes in the query set $\mathcal{Q}^i$. The representations of nodes in $\mathcal{D}^i$ are obtained through the same procedure: treat each node as centric node to retrieve a contextualized subgraph, and then feed the sampled subgraphs to the pretrained GNN encoder, and the centric node embedding can be directly indexed and used to fine-tune the classifier $f_\psi$ by optimizing a naive Cross-Entropy Loss.

\section{Experiments}
In this section, we design experiments to evaluate the proposed framework by comparing with three categories of methods for few-shot node classification: (1) naive supervised pretraining using GNNs, (2) state-of-the-art meta-learning based methods, and (3) by going one step further to compare with contrastive learning methods. For the third category, to our best knowledge, we are the first to implement state-of-the-art self-supervised graph contrastive methods to few-shot node classification and compare them with the proposed graph supervised contrastive learning. %to pretrain the GNN encoder on the base classes and fine-tune a classifier on novel classes, and we compared them with the meta-learning based methods and ours. 
% Now, we introduce experimental settings and then present evaluation results and analysis. 

% Next, we  in terms of training time for convergence. In addition, we show that the proposed method outperforms baselines in clustering unlabeled nodes from novel classes without fine-tuning. This implies that the representations learned by the proposed method elicit decision boundaries that can distinguish samples from different classes with high accuracy. Furthermore, to illustrate the advantage of the proposed loss function, we compare it with three typical candidate loss functions under various settings. Finally, we conduct a comprehensive sensitivity analysis to study and validate the effectiveness of individual components in the proposed framework and investigate their characteristics.
\vspace{-0.2cm}
\subsection{Experimental Settings}
\subsubsection{Evaluation Datasets} 
\begin{table}[]
\vspace{-1.1cm}
\centering
\caption{\label{tab:dataset}Statistics of the commonly used datasets}
\scalebox{1}{
\begin{tabular}{ccccc|cc}
\hline
           & \# Nodes & \# Edges   & \# Features & \# Labels & Base & Novel \\ \hline
CoraFull   & 19,793   & 126,842    & 8,710       & 70       & 42    & 28   \\ \hline
Reddit     & 232,965  & 11,606,919 & 602         & 41       & 24    & 17   \\ \hline
Ogbn-arxiv & 169,343  & 1,166,243  & 128         & 40       & 24    & 16   \\ \hline
\end{tabular}}
\centering
\vspace{-0.5cm}
\end{table}

We conduct our experiments on three widely used graph few-shot learning benchmark datasets where a sufficient number of node classes are available for sampling few-shot node classification tasks: CoraFull \cite{bojchevski2017deep}, Reddit~\cite{hamilton2017inductive}, and Ogbn-arxiv \cite{xu2018powerful}. Their statistics are given in Table \ref{tab:dataset}.

\vspace{-0.4cm}
\subsubsection{Baseline Methods}
In this work, we compare our framework with the following 3 categories of methods.
\label{methods}
\begin{itemize}
    \item Naive supervised pretraining. %A naive GNN encoder. For this experiment, 
  We use GCN \cite{kipf2016semi} as a naive encoder and pretrain it with all nodes from the base classes, and following convention, we fine-tune a single linear layer as the classifier for each few-shot node classification task. 
  Also, we implement an initialization strategy, TFT \cite{dhillon2019baseline} for the classifier by setting its weight as a matrix consisting of concatenated prototype vectors of novel classes. 
%   This method is primarily used for few-shot image classification. We replace its encoder with the same GCN encoder.
  \item State-of-the-art meta-learning methods for few-shot node classification on a single graph: MAML based Meta-GNN \cite{zhou2019meta}, Matching Network \cite{vinyals2016matching} based AMM-GNN \cite{wang2020graph}, and Prototypical Network based GPN~\cite{ding2020graph}. We do not include methods like \cite{yao2020graph,wen2021meta} in the baselines because they require extra auxiliary graph data, nor methods like \cite{lan2020node,liu2021relative} because they have similar performance with the chosen baselines according to their original papers.
  \item State-of-the-art self-supervised pretraining methods on a single graph: MVGRL \cite{hassani2020contrastive}, SUBG-Con \cite{jiao2020sub}, GraphCL \cite{you2020graph}, and GCA \cite{zhu2021graph}. %Following a similar procedure to our framework, 
  These methods pretrain a GNN encoder with nodes from base classes without using the labels. The classifiers are then fine-tuned on novel support nodes and their accuracy is reported on predicting labels for novel query nodes. %The architectures of these classifiers are the same as those for node classification downstream tasks in the original papers.
\end{itemize}

\paragraph{Evaluation Protocol:} To make fair comparison, all the scores reported are accuracy values averaged over 10 random seeds and all the baselines share the same splits of base classes and novel classes as shown in Table~\ref{tab:dataset}.

\subsubsection{Implementation Details:}

In Table \ref{tab:dataset}, the specific data splits are listed for each dataset. For a fair comparison, we adopt the same encoder for all compared methods. Specifically, the graph encoder $g_\theta$ consists of one GCN layer \cite{kipf2016semi} with PReLU activation. The effect of the encoder architecture are further explored in Section~\ref{exp:encoder}. We choose logistic regression as the linear classifier for fine-tuning. The encoder is trained with Adam optimizers whose learning rates are set to be $0.001$ initially with a weight decay of $0.0005$. And the coefficients for computing running averages of gradient and square are set to be $\beta_1 = 0.9$, $\beta_2 = 0.999$. The default values of batch size $B$ and graph temperature parameter $\beta$ are set to 500 and 1.0, respectively. For the baseline methods, we use the default parameters provided in their implementations.

\subsection{Overall Evaluation}
\label{exp:comp}
We present the comparative results between our framework and three categories of baseline methods described earlier. %in Section \ref{methods}. 
It is worth mentioning that, this work is the first to investigate the necessity of episodic meta-learning for Few-shot Node Classification (FNC) problems. For a fair comparison, all methods share the same GCN encoder architecture as the proposed framework. Also, when experimented on each dataset, they share the same random seeds for data split, leading to identical evaluation data. The results are shown in Table \ref{tab:comp}. We summarize our findings next. 

% Pretraining Loss /\\ Meta-training Loss
\begin{table*}[]
\vspace{-0.7cm}
\caption{\label{tab:comp}Comparative Results: three datasets under different N-way K-shot settings}
\scalebox{0.56}{
\begin{tabular}{ccccccccccc}
\hline
\multirow{2}{*}{\textbf{Methods}} & \multirow{2}{*}{\textbf{\begin{tabular}[c]{@{}c@{}}Loss\end{tabular}}} & \multicolumn{3}{c}{\textbf{CoraFull (\%)}}                         & \multicolumn{3}{c}{\textbf{Reddit (\%)}}                           & \multicolumn{3}{c}{\textbf{Ogbn-arxiv (\%)}}     \\ \cline{3-11} 
                                  &                                                                                                           & 10-way 5-shot  & \multicolumn{1}{l}{5-way 5-shot} & 3-way 1-shot   & 10-way 5-shot  & 5-way 5-shot   & \multicolumn{1}{l}{3-way 1-shot} & 10-way 5-shot  & 5-way 5-shot   & 3-way 1-shot   \\ \hline
GCN                               & CEL                                                                                                       & 37.25          & 45.68                            & 43.23          & 36.28          & 44.34          & 39.62                            & 30.83          & 38.40          & 35.41          \\
TFT                               & CEL                                                                                                       & 63.50          & 70.18                            & 66.41          & 59.75          & 69.80          & 58.32                            & 47.68          & 62.25          & 60.48          \\ \hline
Meta-GNN                          & CEL                                                                                                       & 55.23          & 66.25                            & 60.25          & 48.62          & 65.50          & 55.78                            & 41.20          & 58.67          & 55.68          \\
AMM-GNN                           & CEL                                                                                                       & 60.80          & 70.52                            & 65.27          & 53.28          & 67.20          & 55.84                            & 44.33          & 61.02          & 58.64          \\
GPN                               & CEL                                                                                                       & 62.02          & 73.40                            & 67.07          & 59.20          & 69.31          & 60.20                            & 50.58          & 64.12          & 62.20          \\ \hline
GraphCL                           & SimCLR                                                                                                    & 79.68          & 87.35                            & 84.80          & 82.51          & 87.34          & 84.16                            & 57.30          & 67.34          & 62.27          \\
MVGRL                             & DIM                                                                                                       & 81.34          & 88.40                            & 85.03          & 84.78          & 89.65          & 86.29                            & 57.82          & 68.24          & 63.36          \\
SUBG-Con                          & Margin Loss                                                                                             & 80.71          & 88.02                            & 86.26          & 83.70          & 89.55          & 85.57                            & 56.29          & 66.36          & 62.10          \\
GCA                               & SimCLR                                                                                                    & 82.43          & 90.52                            & 87.89          & 85.61          & 90.96          & 87.26                            & 59.03          & 69.53          & 65.49          \\ \hline
\textbf{Ours (NAD)}               & \textbf{G-SupCon}                                                                                         & \textbf{86.13} & \textbf{93.65}                   & 89.42          & 88.52          & 93.36          & 91.70                            & \textbf{62.24} & 73.25          & 71.63          \\
\textbf{Ours (PPR)}               & \textbf{G-SupCon}                                                                                         & 85.34          & 93.62                            & \textbf{90.56} & \textbf{89.10} & \textbf{94.12} & \textbf{92.31}                   & 61.85          & \textbf{73.65} & \textbf{72.90} \\ \hline
\end{tabular}}
\vspace{-.4cm}
\end{table*}

\textbf{Necessity} of employing episodic meta-learning style methods for graph few-shot learning. First and foremost, we find that almost all the contrastive pretraining based methods outperform the existing meta-learning based FNC algorithms. Even the most straightforward one, TFT, which only leverages a simple initialization to the separate classifier, can produce comparable scores to the best meta-learning based FNC method, GPN. We have shown that through appropriate pretraining, including self-supervised and supervised training, adding a simple linear classifier can outperform the existing meta-learning based framework by a significant margin.

\textbf{Effectiveness} of our framework to learn discriminative node embeddings for FNC problems. Compared with other existing meta-learning based methods, our framework outperforms them by a large margin under different settings. We attribute this to the following facets: (1) The effectiveness of the proposed node-connectivity-based sampling strategy that can provide highly correlated context-specific information for the centric node by considering both global and local information. Besides, NAD and PPR can provide similar outcomes. NAD can perform better on graphs with a higher average degree; and (2) The supervised contrastive learning loss function G-SupCon can utilize label information to further enforce the GNN encoder to generate more discriminative representations by minimizing the distances among nodes from the same classes while segregating nodes from different classes in the representation space. 

\textbf{Robustness} to various $N$-way $K$-shot settings. Similar to the meta-learning based FNC methods, the performance of contrastive pretraining based methods also degrades when the $K$ decreases or $N$ increases. However, from the results shown in Table \ref{tab:comp}, we find that those methods are more robust to settings with decreasing $K$ and increasing $N$. This means that the encoder can better extrapolate to novel classes by generating more discriminative node representations. To a large extent, the degradation lies in the less accurate classifier due to fewer training nodes. Learning to better measure the classifier under scenarios with extremely scarce support nodes (very small $K$) is also worth further research.

\subsection{Further Experiments}

To further evaluate our framework, we conduct more experiments next. %Without further clarification, 
The default setting for the following experiments is $5$-way $5$-shot, where we set $\beta = 1.0$, $B = 500$, and use a single GCN layer as the encoder.

\vspace{-0.5cm}
\subsubsection{Efficiency Study.} 
As discussed in Section \ref{fast}, our method is scalable and has much less convergence time because we construct a GNN encoder for the sampled subgraph rather than the whole graph. Also, Section \ref{exp:comp} shows that performance is not sacrificed for achieving scalability. On the contrary, due to the effective sampling strategy, only fine-grained data are fed into the encoder, resulting in a considerable boost in accuracy. We show the actual time consumption for a single run of training of our model and typical meta-learning and pretraining baselines. The experiment is conducted on a single RTX 3090 GPU. In Table~\ref{tab:time}, we show that our method can achieve excellent performance and consume much less training time. Note that here we only consider the training time, excluding the time for computing the node connectivity scores $\mathbf{S}$ which can be pre-calculated.

\begin{table}[]
\vspace{-0.9cm}
\centering
\caption{\label{tab:time} Training time of GPN, GCA, and Ours (proposed method)}
\begin{tabular}{cccc}
\hline
\multirow{1}{*}{Methods}
                         & \textbf{CoraFull} & \textbf{Reddit} & \textbf{Ogbn-arxiv} \\ \hline
GPN                      & 1352s             & 3248s           & 2562s               \\ \hline
GCA                      & 584s              & 3651s           & 3237s               \\ \hline
\textbf{Ours}            & \textbf{10s}      & \textbf{21s}    & \textbf{54s}        \\ \hline
\end{tabular}
\centering
\vspace{-1.1cm}
\end{table}
%  & \multicolumn{3}{c}{Datasets}                              \\ \cline{2-4} 

\subsubsection{Representation Clustering.}
\begin{figure}[h]
  \centering
  \scalebox{0.8}{
  \includegraphics[width=\linewidth]{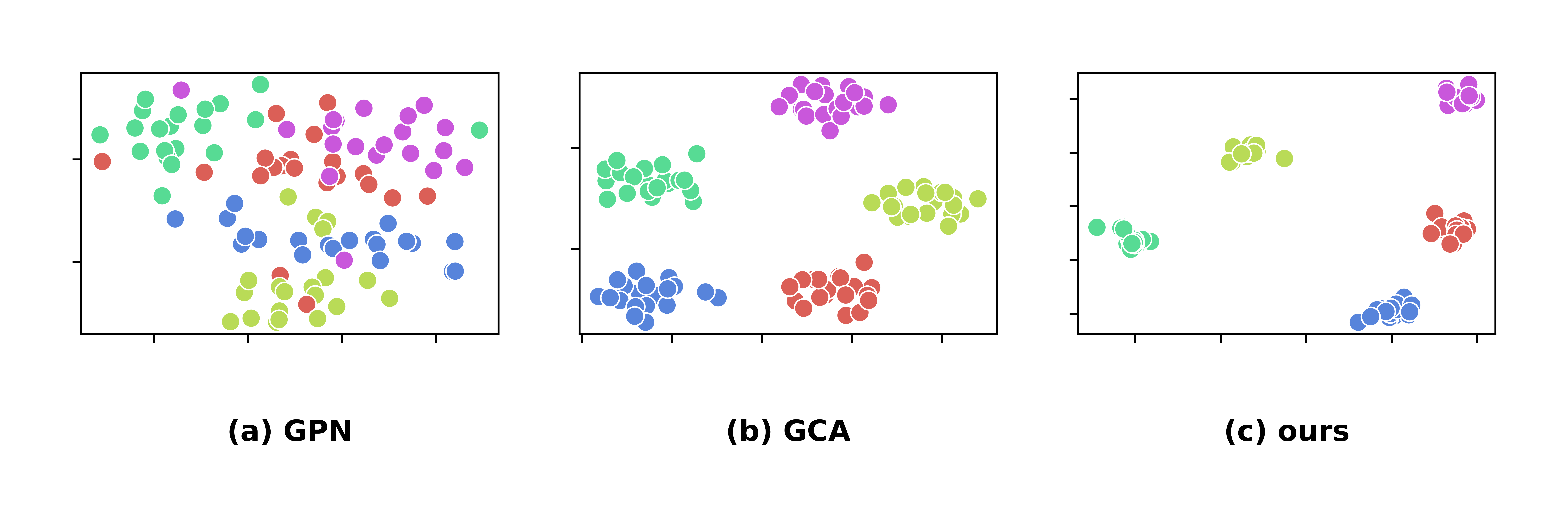}}
  \vspace{-0.6cm}
  \caption{t-SNE embedding visualization on CoraFull data: (a) GPN (b) GCA (C) Ours}
  \label{fig:cluster}
  \vspace{-0.4cm}
\end{figure}

This experiment is designed to demonstrate the high-quality representation generated by the encoder in the proposed framework. We show the clustering results of the representation of the query nodes in 5 randomly sampled novel classes in Table~\ref{tab:cluster} and visualize the embedding in Figure~\ref{fig:cluster} (without fine-tuning on support set). It can be observed that an encoder trained by the proposed pretraining strategy possesses a stronger extrapolation ability to generate highly discriminative boundaries for unseen novel classes.

\begin{table}[]
\centering
\vspace{-0.2cm}
\caption{\label{tab:cluster}Performance on novel classes query node embedding clustering, reported in Normalized Mutual Information (\textbf{NMI}) and Adjusted Rand Index (\textbf{ARI})}
\scalebox{1}{
\begin{tabular}{ccccccc}
\hline
\multirow{2}{*}{Methods} & \multicolumn{2}{c}{\textbf{CoraFull}} & \multicolumn{2}{c}{\textbf{Reddit}} & \multicolumn{2}{c}{\textbf{Ogbn-arxiv}} \\ \cline{2-7} 
                         & NMI                  & ARI                 & NMI                 & ARI                & NMI                   & ARI                  \\ \hline
GPN                      & 0.5134               & 0.4327              & 0.3690              & 0.3115             & 0.2905                & 0.2235               \\
GCA                      & 0.7531               & 0.7351              & 0.7824              & 0.7756             & 0.3786                & 0.3219               \\ \hline
\textbf{Ours}            & \textbf{0.8567}      & \textbf{0.8229}     & \textbf{0.8890}     & \textbf{0.8720}    & \textbf{0.5280}       & \textbf{0.4485}      \\ \hline
\end{tabular}}
\centering
\vspace{-0.5cm}
\end{table}

\vspace{-0.8cm}
\subsection{Ablation \& Parameter Analysis}

In this section we study how sensitive the proposed framework is to the design choice of its components. In particular, we consider the architecture of encoder $g_\theta$, graph temperature parameter $\beta$, and batch size $B$. 

\noindent\textbf{Analysis on Encoder Architecture.}
Our framework is independent of the encoder architecture. We evaluate our framework with three widely-used GNNs: GCN \cite{kipf2016semi}, GAT \cite{velivckovic2017graph}, and GIN \cite{xu2018powerful}, as shown in Table \ref{tab:encoder}. The difference between encoder architectures is insignificant, so we choose GCN as the default encoder.

\noindent\textbf{Analysis on Loss Function.}
%\label{exp:abl}
To better demonstrate the effectiveness of our G-SubCon loss, we design an experiment to compare different loss functions. The SimCLR loss does not consider the label information, so the Balance Sampling (BS) strategy we proposed in Section \ref{BS} is not feasible for it. But in order to explicitly show the influence from the loss function, we also show the result of SimCLR with the same sampled data splits from BS as our G-SubCon. We list the results from all the candidate loss functions under both settings in Table \ref{tab:loss}. It can be observable that all the losses suffer from the class imbalance issue, and the simple BS scheme can improve performance. Also, our proposed G-SupCon outperforms all others even with identical data splits. In short, both the BS scheme and the G-SupCon loss function are effective in terms of accuracy on few-shot node classification tasks.

\label{exp:encoder}
\begin{table}[]
\vspace{-0.7cm}
\centering
\caption{\label{tab:encoder}Results of our model with different encoders}
\scalebox{1}{
\begin{tabular}{cccc}
\hline
Encoder & \textbf{CoraFull (\%)} & \textbf{Reddit (\%)} & \textbf{Ogbn-arxiv (\%)} \\ \hline
GCN     & \textbf{93.62}         & \textbf{94.12}       & 73.65                    \\ 
GAT     & 92.05                  & 94.04                & 73.46                    \\ 
GIN     & 93.25                  & 93.86                & \textbf{74.10}           \\ \hline
\end{tabular}}
\centering
\vspace{-1.0cm}
\end{table}

\subsubsection{Analysis on Graph Temperature Parameter and Batch Size.}
\begin{figure}[]
  \vspace{-1.2cm}
  \centering
  \scalebox{0.7}{
  \includegraphics[width=\linewidth]{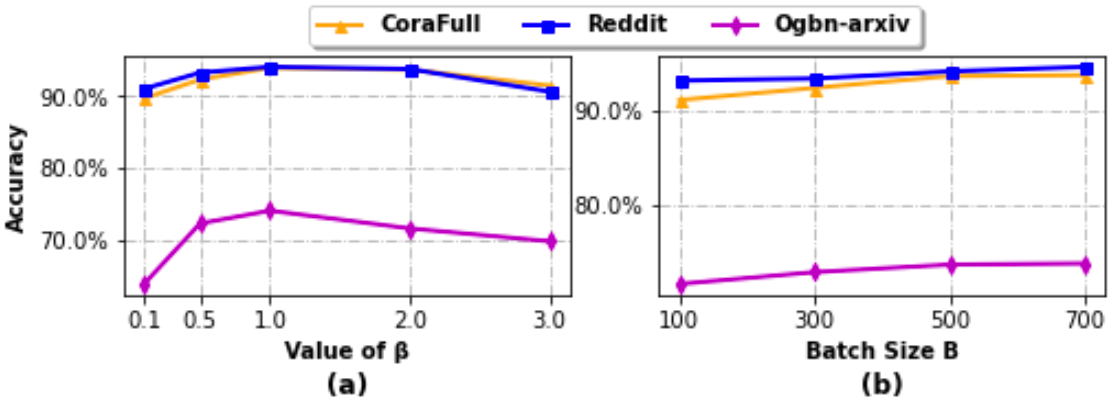}}
  \vspace{-0.2cm}
  \caption{(a) Accuracy vs Graph temperature parameter $\beta$ (b) Accuracy vs Batch size $B$ on the three datasets}
  \label{fig:batch}
  \vspace{-1cm}
\end{figure}

As presented in Figure \ref{fig:batch} (a), we test the sensitivity of our framework regarding the graph temperature parameter $\beta$ and batch size $B$. Observably, our framework is not that sensitive to these two hyperparameters. The best value of $\beta$ is $1.0$ and we set it as default. Generally speaking, the larger batch size can produce higher scores because more contrast can be made in each batch. We choose $500$ as the default batch size for computational efficiency and its decent performance.

\begin{table}[]
\vspace{-0.8cm}
\centering
\caption{\label{tab:loss}Results of our model trained with different loss functions}
\scalebox{0.8}{
\begin{tabular}{ccccc}
\hline
Loss Function     & BS           & \textbf{CoraFull (\%)} & \textbf{Reddit (\%)} & \textbf{Ogbn-arxiv (\%)} \\ \hline
Cross Entropy     & No           & 70.18                  & 69.80                & 62.25                    \\
Cross Entropy     & Yes          & 73.86                  & 74.08                & 63.67                    \\ \hline
SimCLR            & No           & 87.54                  & 89.60                & 66.34                    \\
SimCLR            & Yes          & 89.48                  & 92.88                & 69.98                    \\ \hline
\textbf{G-SupCon} & \textbf{Yes} & \textbf{93.62}         & \textbf{94.12}       & \textbf{73.65}           \\ \hline
\end{tabular}}
\centering
\vspace{-0cm}
\end{table}

\section{Related Work}

\textbf{Few-shot Node Classification.}
Graph Neural Networks (GNNs) \cite{kipf2016semi,velivckovic2017graph,xu2018powerful} are a family of deep neural network models for graph-structured data, which exploits recurrent neighborhood aggregation to preserve the graph structure information and transform the node attributes simultaneously. Recently, increasing attention has been paid to few-shot node classification problems, episodic meta-learning~\cite{finn2017model} has become the most dominant paradigm. It trains the GNN encoders by explicitly emulating the test environment for few-shot learning~\cite{finn2017model}, where the encoders are expected to gain the adaptability to extrapolate onto new domains. Meta-GNN \cite{zhou2019meta} applies MAML \cite{finn2017model} to learn directions for optimization with limited labels. AMM-GNN \cite{wang2020graph} deploys Matching Network \cite{vinyals2016matching} to learn transferable metric among different meta-tasks.  GPN~\cite{ding2020graph} adopts Prototypical Networks \cite{snell2017prototypical} to make the classification based on the distance between the node feature and the prototypes. MetaTNE~\cite{lan2020node} and RALE \cite{liu2021relative} also use episodic meta-learning to enhance the adaptability of the learned GNN encoder and achieve similar results. Furthermore, HAG-Meta~\cite{tan2022graph} extends the problem to incremental learning setting. Recently, people in the image domain argue that the reason for the fast adaptation in the existing works lies in feature reuse rather than those complicated mate-learning algorithms~\cite{dhillon2019baseline,tian2020rethinking}. In other words, with a carefully pretrained encoder, decent performance can be obtained through direct fine-tuning a simple classifier on the target domain. Since then, various pretraining strategies \cite{tian2020rethinking,liu2021learning} have been put forward to tackle the few-shot image classification problem. However, no research has been done in the graph domain with its crucial distinction from images that nodes in a graph are not i.i.d. data. Their interactive relationships are reflected by both the topological and semantic information. Our work here is the first attempt to bridge the gap by developing a novel graph supervised contrastive learning for few-shot node classification.

% G-Meta \cite{huang2020graph}, GFL-KT \cite{yao2020graph} and MI-GNN~\cite{wen2021meta} use meta-learning to transfer knowledge when auxiliary graphs exist, and

\textbf{Graph Contrastive Learning.}
Contrastive learning has become popular representation learning paradigm in image \cite{chen2020simple}, text \cite{wang2021cline}, and graph \cite{hassani2020contrastive,zhu2021graph} domains. Starting from the self-supervised setting, contrastive learning methods are proved to learn discriminative representation by contrasting a predefined distance between positive and negative samples. Usually, those samples are augmented through some heuristic transformations from original data. Specifically, in the graph domain, the transformations can be categorized into the following types: (1) graph structure based augmentation, e.g., randomly drop edges or nodes \cite{hassani2020contrastive,zhu2021graph}, randomly sample subgraphs \cite{jiao2020sub}. (2) graph feature based augmentation, e.g., randomly mask or perturb attributes of nodes or edges \cite{tong2021directed}. \cite{zhu2021graph} further improves those augmentations by adding masks according to feature importance. Some works \cite{you2021graph,suresh2021adversarial,you2022bringing} try to explore different ways to automatically generate augmented views. 
% To further enhance contrastive learning, other works investigate the training procedures. For example, \cite{zhao2021graph} uses clustering to mine true-negative pairs to ensure the negative contrast. \cite{chu2021cuco} proposes to use Curriculum Learning \cite{bengio2009curriculum} to rank the hardship of negative samples and enforce the GNN model to learn from simple to hard. 
% Another interesting point for graph contrastive learning lies in different contrastive scales \cite{liu2021graph}. Instead of contrast between the same scale level, like node-level and graph-level \cite{hassani2020contrastive,you2020graph}, some works deploy a cross-scale contrast mechanism, where nodes are contrast with graphs~\cite{zhu2020deep,jiao2020sub,zhu2021graph}. 
% To accommodate different real-world applications, people have researched contrastive learning on special types of graphs, such as user-item interaction bipartite graphs \cite{wu2021self}, item-session transition graphs \cite{xia2021self1,xia2021self2}, heterogeneous graphs \cite{wang2021self,jiang2021contrastive}, molecular graphs \cite{sun2021mocl,zhao2021csgnn}, knowledge graph \cite{xu2021kge}, signed graph \cite{shu2021sgcl}.
Recently, for image \cite{khosla2020supervised} and text~\cite{gunel2020supervised} domains, people notice that by injecting label information to the contrastive loss to compact or enlarge the distances of augmentations of instances within the same or different classes, supervised contrastive loss outperforms the original unsupervised version and even cross-entropy loss in the setting of transfer learning, by providing highly discriminative representation learned from texts or images. However, no existing work has focused on its extrapolation ability for graphs, especially, under an extremer few-shot situation. 

\vspace{-0.25cm}
\section{Conclusion, Limitations and Outlook}
\vspace{-0.2cm}
In this work, we question the fundamental question whether episodic meta-learning is necessary for few-shot node classification. To answer the question, we propose a graph supervised contrastive learning tailored for the few-shot node classification problem and demonstrate its superb adaptability to extrapolate onto novel classes by fine-tuning a simple linear classifier. Through extensive experiments on benchmark datasets, we demonstrate that our framework can surpass episodic meta-learning methods for few-shot node classification in terms of both accuracy and efficiency. Therefore, this work offers the answer: episodic meta-learning is not a must for few-shot node classification.

Due to limited space, limitations of our work need to be acknowledged. (1)~\textit{Limited strategy consideration.} To pretrain the GNN encoder on base classes, we only consider naive supervised training and graph contrastive training. There are many other pretraining strategy that worth further investigation (e.g.~\cite{hu2020strategies}).
(2)~\textit{Lack of theoretical justification.} Our work mainly presents empirical studies which may throw up many questions in need of further theoretical justification, for instance, to what magnitude contrastive pretraining surpasses meta-learning and the reason behind it. 

We hope our work will shed new light on few-shot node classification tasks. There are many promising directions worth further research. For example, when base classes also have very limited labeled nodes or even no label at all, from Table \ref{exp:comp}, we can see that self-supervised pretraining can also help improve the performance. So it would be interesting to investigate the pretraining strategy under semi-supervised or unsupervised settings. In addition, since the pretraining phase may involve extra noise through data sampling or augmentation, methods to calibrate the learned embedding or refine the obtained prototypes are also potential directions for research.

\vspace{-0.1cm}
\section{Acknowledgments}
\noindent This work is partially supported by Army Research Office (ARO)
W911NF2110030 and Army Research Lab (ARL) W911NF2020124.

%
% ---- Bibliography ----
%
% BibTeX users should specify bibliography style 'splncs04'.
% References will then be sorted and formatted in the correct style.
%
\bibliographystyle{splncs04}
\bibliography{ecml22}
%
% \begin{thebibliography}{8}
% \bibitem{ref_article1}
% Author, F.: Article title. Journal \textbf{2}(5), 99--110 (2016)

% \bibitem{ref_lncs1}
% Author, F., Author, S.: Title of a proceedings paper. In: Editor,
% F., Editor, S. (eds.) CONFERENCE 2016, LNCS, vol. 9999, pp. 1--13.
% Springer, Heidelberg (2016). \doi{10.10007/1234567890}

% \bibitem{ref_book1}
% Author, F., Author, S., Author, T.: Book title. 2nd edn. Publisher,
% Location (1999)

% \bibitem{ref_proc1}
% Author, A.-B.: Contribution title. In: 9th International Proceedings
% on Proceedings, pp. 1--2. Publisher, Location (2010)

% \bibitem{ref_url1}
% LNCS Homepage, \url{http://www.springer.com/lncs}. Last accessed 4
% Oct 2017
% \end{thebibliography}
\end{document}